\documentclass[conference]{IEEEtran}

\usepackage{cite}
\usepackage{amsmath,amssymb,amsfonts}
\usepackage{algorithmic}
\usepackage{graphicx}
\usepackage{textcomp}
\usepackage{xcolor}
\usepackage{booktabs}
\usepackage{multirow}
\usepackage{tabularx}
\usepackage{array}
\usepackage{microtype}
\usepackage[hidelinks]{hyperref}

\begin{document}

\title{SlugTrails: An Egocentric Benchmark for Floor Plan Localization in Large Buildings}

\author{\IEEEauthorblockN{Yunqian Cheng}
\IEEEauthorblockA{\textit{Department of Computer Science and Engineering}\\
\textit{University of California, Santa Cruz}\\
Santa Cruz, CA, USA\\
ychen827@ucsc.edu}
\and
\IEEEauthorblockN{Roberto Manduchi}
\IEEEauthorblockA{\textit{Department of Computer Science and Engineering}\\
\textit{University of California, Santa Cruz}\\
Santa Cruz, CA, USA\\
manduchi@ucsc.edu}}

\maketitle

\begin{abstract}
Floor-plan-based indoor visual localization enables infrastructure-free positioning, but most methods are developed and evaluated in small residential environments unlike the large public buildings of real deployment. We introduce \textbf{SlugTrails}, a floor plan localization benchmark for large indoor spaces under realistic egocentric sensing: $30$\,Hz Aria glasses recordings across three campus buildings and six floors ($22\,089$\,m$^{2}$ of floor plan outline), CAD-derived floor plans with semantic classes and circulation space masks, and trajectories aligned into the floor plan frame using laser-surveyed anchors. One protocol covers three practical ways of gathering geometry under a limited field of view---a single walking frame, a stationary multi-view sweep, and a walking stream with odometry---so methods designed for different regimes are compared on the same buildings and ground truth. Evaluating five representative geometric and learned systems under their native sensing configurations, we find that stock checkpoints (official released weights) are near zero on SlugTrails (at most $0.004$ R@1m30$^{\circ}$ on walking single frames), while fine-tuning on SlugTrails improves every trainable family on all three tasks (e.g., F$^{3}$Loc $0.0\rightarrow 0.141$ single-frame and $0.03\rightarrow 0.66$ sequential), with gains compounding as observations accumulate. The same fine-tuned weights also improve cross-dataset generalization on LaMAR with no LaMAR training (sequential R@1m $0.048\rightarrow 0.143$ for F$^{3}$Loc and $0.063\rightarrow 0.127$ for UnLoc), whereas train-from-scratch on SlugTrails alone stays far below fine-tuning from stock weights---evidence that floor plan localization is currently limited by indoor data rather than by architecture. We release the dataset, protocols, and tools at \url{https://github.com/Head-inthe-Cloud/SlugTrails}.
\end{abstract}



\section{Introduction}
\label{sec:intro}

Indoor visual localization estimates device pose from visual observations and structural priors. Floor-plan-based localization is attractive because architectural floor plans are widely available and stable, enabling infrastructure-free positioning without an environment-specific image database, a pre-built 3D reconstruction, or dedicated sensing infrastructure.

Existing floor plan localization benchmarks are dominated by apartments and houses with compact search spaces and distinctive local geometry~\cite{Zheng2019Structured3DAL,Shen2020iGibson1A,Ramakrishnan2021HabitatMatterport3D,Cruz2021ZillowID}. Real deployments often occur in large public buildings---academic complexes, hospitals, offices, and transportation facilities---where practical navigation concentrates in \emph{circulation spaces}: corridors, lobbies, stairwells, and connecting passages. These regions frequently contain repeated structural elements, long visually similar corridors, and few globally distinctive cues, so multiple distant poses can be geometrically consistent with the same local observation. Residential-scale evaluation therefore does not reliably predict large-building behavior. A useful benchmark must support both evaluation under realistic large-building conditions and improvement of existing algorithms---for this, we provide egocentric data aligned to the floor plan for diagnosis, fine-tuning, and failure analysis.

\begin{figure*}[t]
\centering
\includegraphics[width=0.8\textwidth]{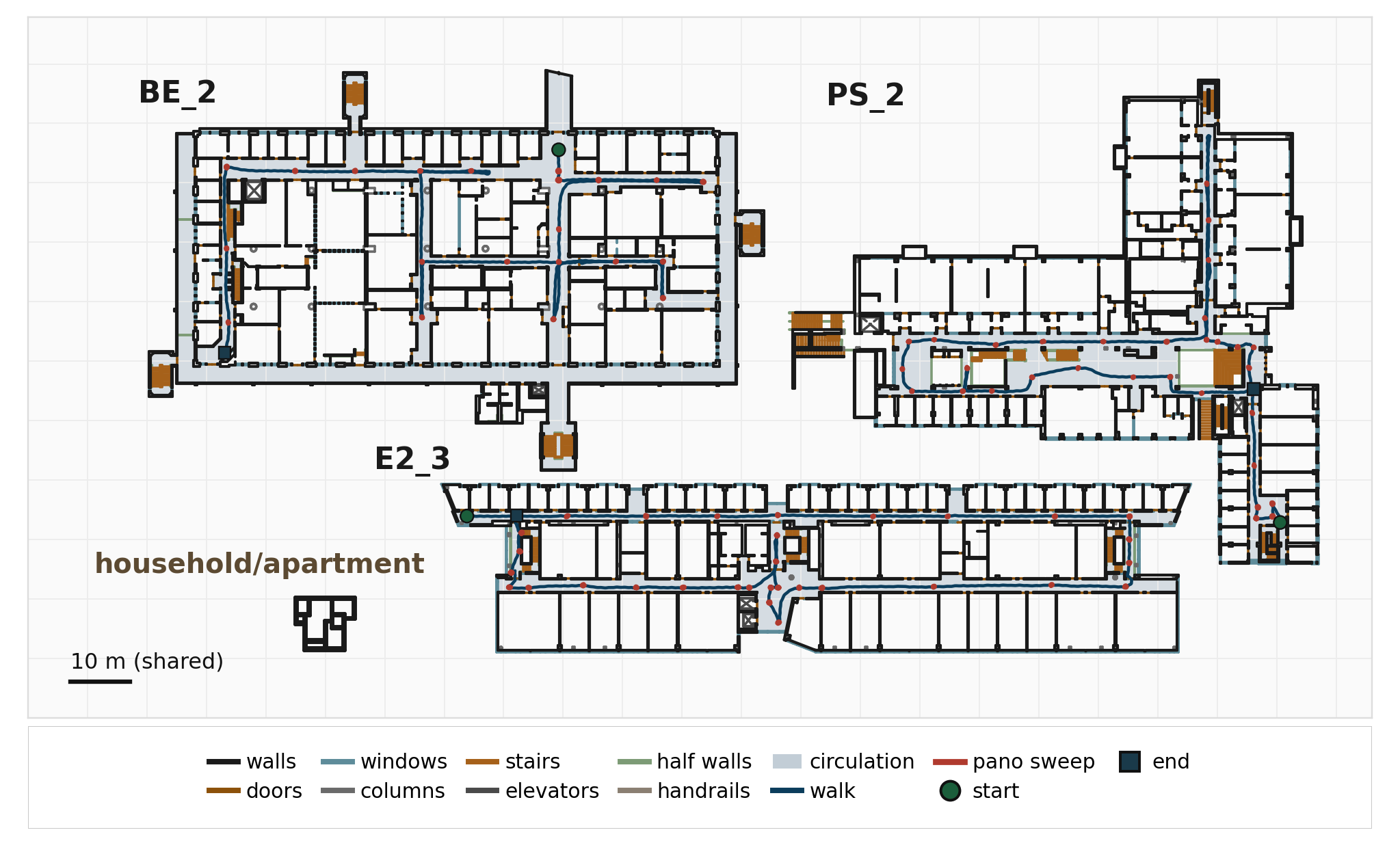}
\caption{SlugTrails large building floor plans and a household-scale Structured3D~\cite{Zheng2019Structured3DAL} floor plan in a shared metric frame (meters).}
\label{fig:floor-plans}
\end{figure*}

We introduce \textbf{SlugTrails}, a benchmark for floor plan localization in large indoor spaces under realistic egocentric sensing (Figs.~\ref{fig:floor-plans} and~\ref{fig:data-collection}). SlugTrails contains recordings aligned to the floor plan collected with Aria glasses~\cite{Somasundaram2023ProjectAA} across multi-floor campus buildings, with metrically verified architectural floor plans and trajectory annotations in the floor plan frame. We evaluate floor plan localization methods including PALMS+~\cite{Cheng2025PALMSMI}, F$^{3}$Loc~\cite{Chen2024F3LocFA}, UnLoc~\cite{West2025UnLocLD}, DisCo-FLoc~\cite{Zhong2026DisCoFLocSF}, and SemRayLoc~\cite{Grader2025SuperchargingFL}, each under its native sensing configuration, and compare every trainable method against stock, SlugTrails fine-tuned, and train-from-scratch variants of itself.

\begin{figure}[t]
\centering
\includegraphics[width=\columnwidth]{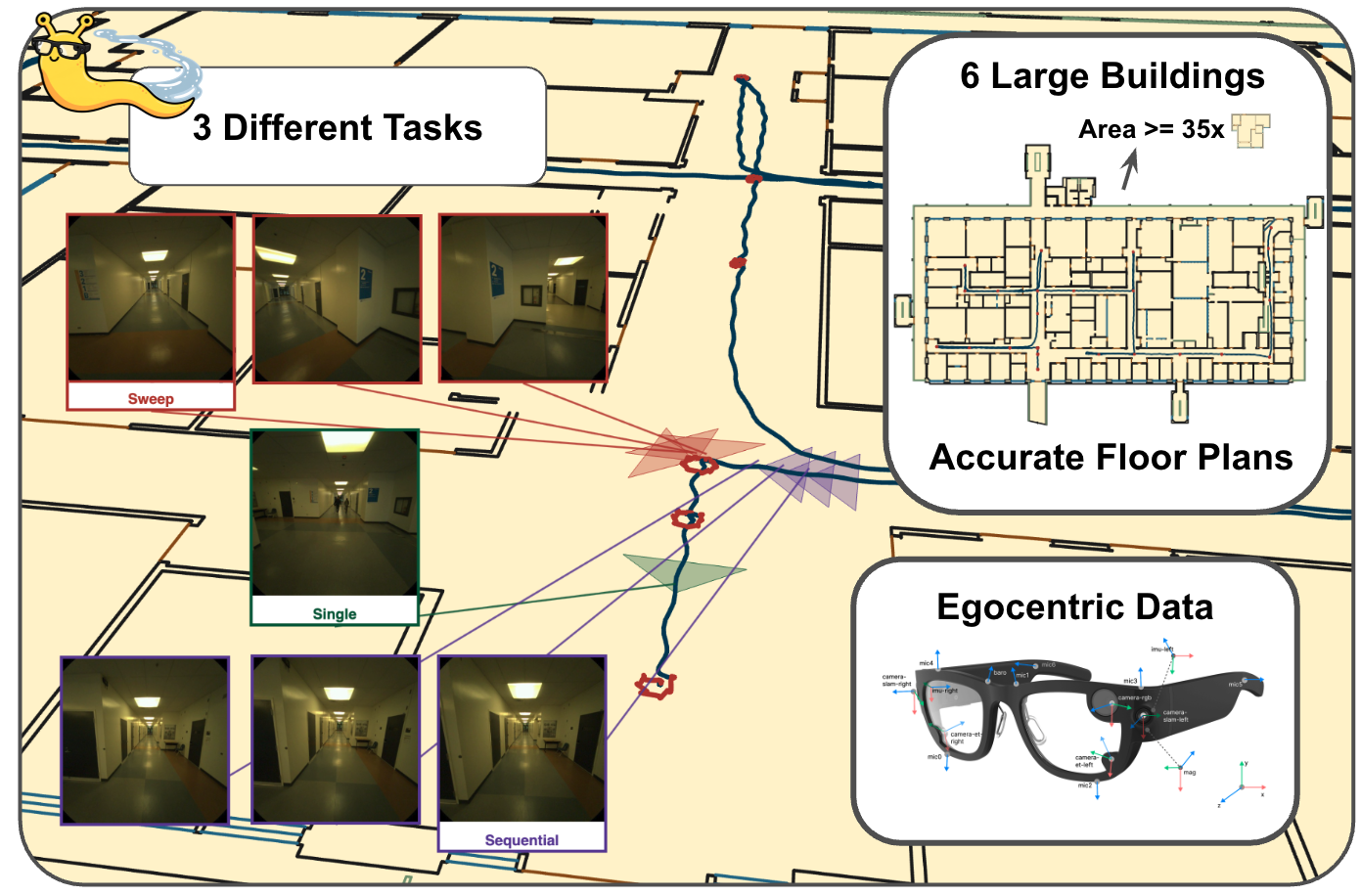}
\caption{The SlugTrails benchmark dataset}
\label{fig:thumbnail}
\end{figure}

Our contributions are:
\begin{enumerate}
    \item \textbf{SlugTrails, a benchmark for realistic large-scale floor plan localization.} Egocentric recordings aligned to the floor plan in large public buildings with single-frame, multi-view sweep, and sequential localization tasks.
    \item \textbf{An evaluation and adaptation study of existing localization systems.} We evaluate representative methods in their official released configurations and compare each to stock, SlugTrails fine-tuned, and train-from-scratch variants of itself. Fine-tuning on SlugTrails improves every trainable system we test, with large gains for several methods under their native observation regimes. Evaluating the same fine-tuned checkpoints on LaMAR~\cite{Sarlin2022LaMARBL} (no LaMAR training) shows that these improvements generalize across datasets.
\end{enumerate}


\section{Related Work}
\label{sec:related}
\subsection{Floor-Plan-Based Indoor Visual Localization}
Floor plan localization offers an infrastructure-free alternative to image or 3D maps by matching egocentric observations to architectural layout.
A large recent cluster of methods---and our main evaluation focus---combines \emph{learned} geometric observations with \emph{geometric} matching to the floor plan and probabilistic pose inference: F$^{3}$Loc~\cite{Chen2024F3LocFA}, UnLoc~\cite{West2025UnLocLD}, DisCo-FLoc~\cite{Zhong2026DisCoFLocSF}, and SemRayLoc~\cite{Grader2025SuperchargingFL} predict horizontal depth or semantic rays and score them against a distance field or occupancy representation, optionally accumulating sequential evidence with a histogram filter. These methods also differ in FOV, ray or layout discretization, depth range, temporal inference, and training data.
Other approaches differ in how they bridge image and floor plan.
Embedding-based methods such as LaLaLoc and LaLaLoc++~\cite{HowardJenkins2021LaLaLocLL,HowardJenkins2022LaLaLocGF} localize by matching learned image and map features in a shared latent space.
Training-free systems recover layout by lifting RGB to 3D with pretrained deep learning models and apply different geometric matchers: PALMS+~\cite{Cheng2025PALMSMI} builds a layout from monocular depth estimation models and convolves it with the floor plan (with CES constraints), while Z-FLoc~\cite{Umemura2026ZFLocZF} utilizes off-the-shelf 3D reconstruction models to reconstruct a bird's-eye view and aligns line and circle primitives from floor plans.

A commonly discussed limitation from the mentioned works is the scarcity of large-building evaluation data with architectural floor plans: most public benchmarks remain household-scale, and when methods are tested in large interiors it is usually on ad hoc custom captures or partial reuse of datasets without floor plans, without a common testing ground.
In those settings, authors report failure modes that are mild at household-scale but severe in large buildings---long corridors, repeated geometry, and symmetric intersections make several distant poses globally plausible for the same local observation.
This motivates a large indoor dataset with metrically aligned floor plans, so future floor plan localization methods can be developed and compared under one protocol to tackle these realistic failure modes.

\subsection{Large-Scale and Egocentric Indoor Localization Datasets}
Popular indoor datasets such as iGibson~\cite{Shen2020iGibson1A}, HM3D~\cite{Ramakrishnan2021HabitatMatterport3D}, ZInD~\cite{Cruz2021ZillowID}, and Structured3D~\cite{Zheng2019Structured3DAL} provide high-quality geometry but are household-scale. Larger localization datasets such as InLoc~\cite{Taira2019InLocIV} and LaMAR~\cite{Sarlin2022LaMARBL} contain realistic public indoor spaces and accurate trajectories, but were not designed as floor plan localization benchmarks. So the floor plans are either not public or not well prepared to be ingested by floor plan localization methods. 

On the other hand, there has been a rise in popularity of wearable devices for the convenience they provide, making them ideal devices for vision-based localization and navigation. Egocentric datasets such as Ego4D~\cite{Grauman2021Ego4DAT} and Aria Digital Twin~\cite{Pan2023AriaDT} provide rich wearable observations, but they lack metrically aligned architectural floor plans. SlugTrails addresses this gap and provides egocentric data collected using Aria glasses.


\section{The SlugTrails Dataset}
\label{sec:dataset}

\subsection{Overview}
SlugTrails is a benchmark for floor plan localization in large indoor spaces under realistic egocentric sensing.
It contains 30\,Hz egocentric RGB video, raw Meta Project Aria Machine Perception Services (MPS) trajectories and poses aligned to the floor plan, and architectural floor plans derived from CAD.
Floor plans include structural walls and semantic classes (doors, windows, columns, stairs, elevators, etc.), together with manually labeled building-outline and circulation space masks (corridors, lobbies, stairwells, and connecting passages).
The release covers three buildings and six floors.
Outline area is $3.0$--$4.3$\,k\,m$^{2}$ per floor ($22\,089$\,m$^{2}$ in total); circulation is $0.62$--$1.37$\,k\,m$^{2}$ per floor ($5317$\,m$^{2}$ in total).
Aligned trajectories total $1.80$\,km including panoramic sweeps and $1.57$\,km of walking between them.
Each floor is therefore about $35$--$50\times$ the outline area of a typical Structured3D apartment ($\sim$87\,m$^{2}$; Fig.~\ref{fig:floor-plans}).
Wall-to-wall laser measurements (Leica DISTO~X4) over twelve hallway spans agree with the floor plans to $1.4$\,cm median absolute discrepancy.

\begin{figure}[t]
\centering
\includegraphics[width=\columnwidth]{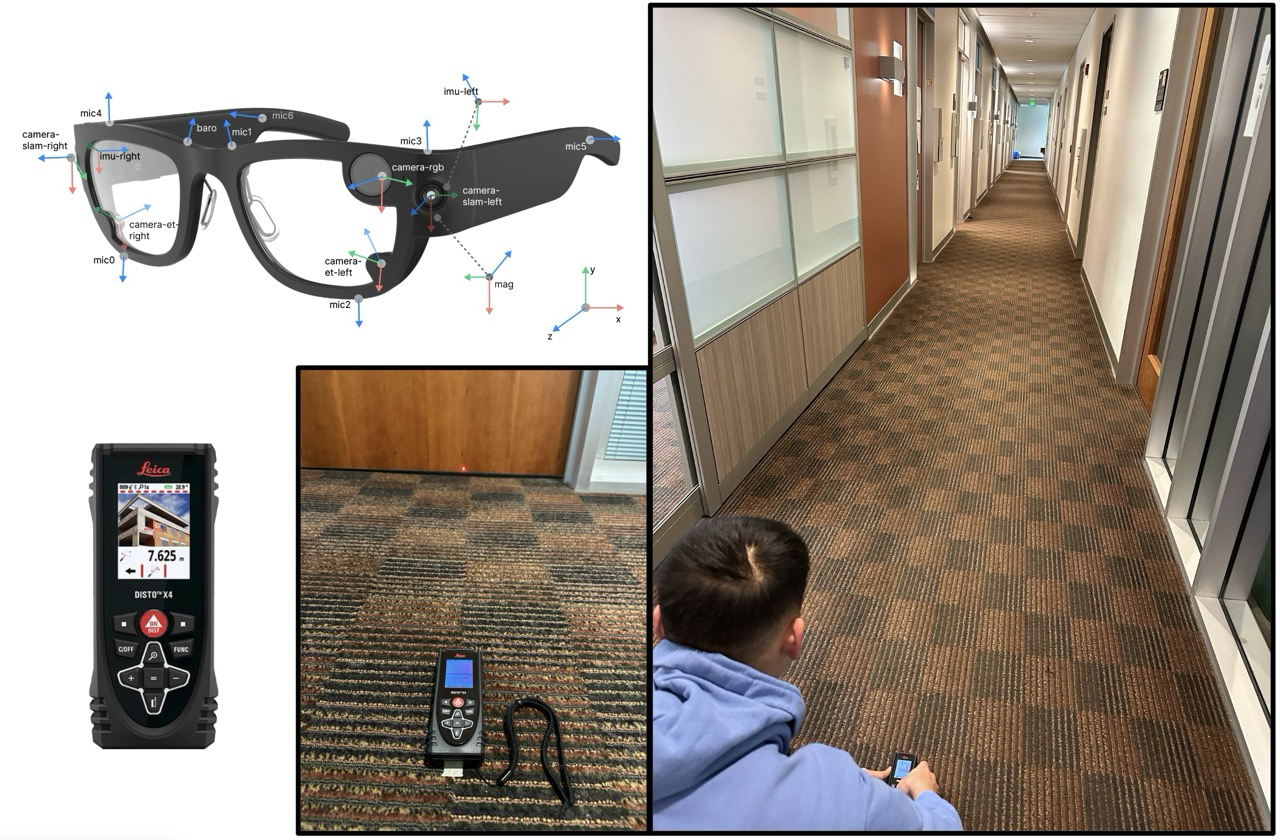}
\caption{SlugTrails data collection. \textbf{Top left:} Aria glasses for egocentric data capture. \textbf{Bottom left:} anchoring a marker using a laser distance measurer (Leica DISTO~X4). \textbf{Right:} researcher measuring hallway length to verify floor plans.}
\label{fig:data-collection}
\end{figure}

\subsection{Data Collection}
\label{sec:data_collection}
Three collectors record with Aria glasses in public spaces of the same buildings.
Temporary floor markers are placed in advance; each collector walks to a marker, performs a stationary $360^{\circ}$ panoramic sweep, then walks to the next marker.
Marker locations are surveyed with the laser distance measurer relative to surrounding walls and registered to the floor plan.
Walking routes are designed to cover most of each floor.
Capture takes place during normal school operation, so sequences include dynamic clutter such as people.
The public release is restricted to public spaces and anonymized with EgoBlur~\cite{Raina2023EgoBlurRI} followed by manual review to remove personal identification information. The data collection process is illustrated in Fig.~\ref{fig:data-collection}.

\subsection{Trajectories and Maps}
\begin{figure}[t]
\centering
\includegraphics[width=\columnwidth]{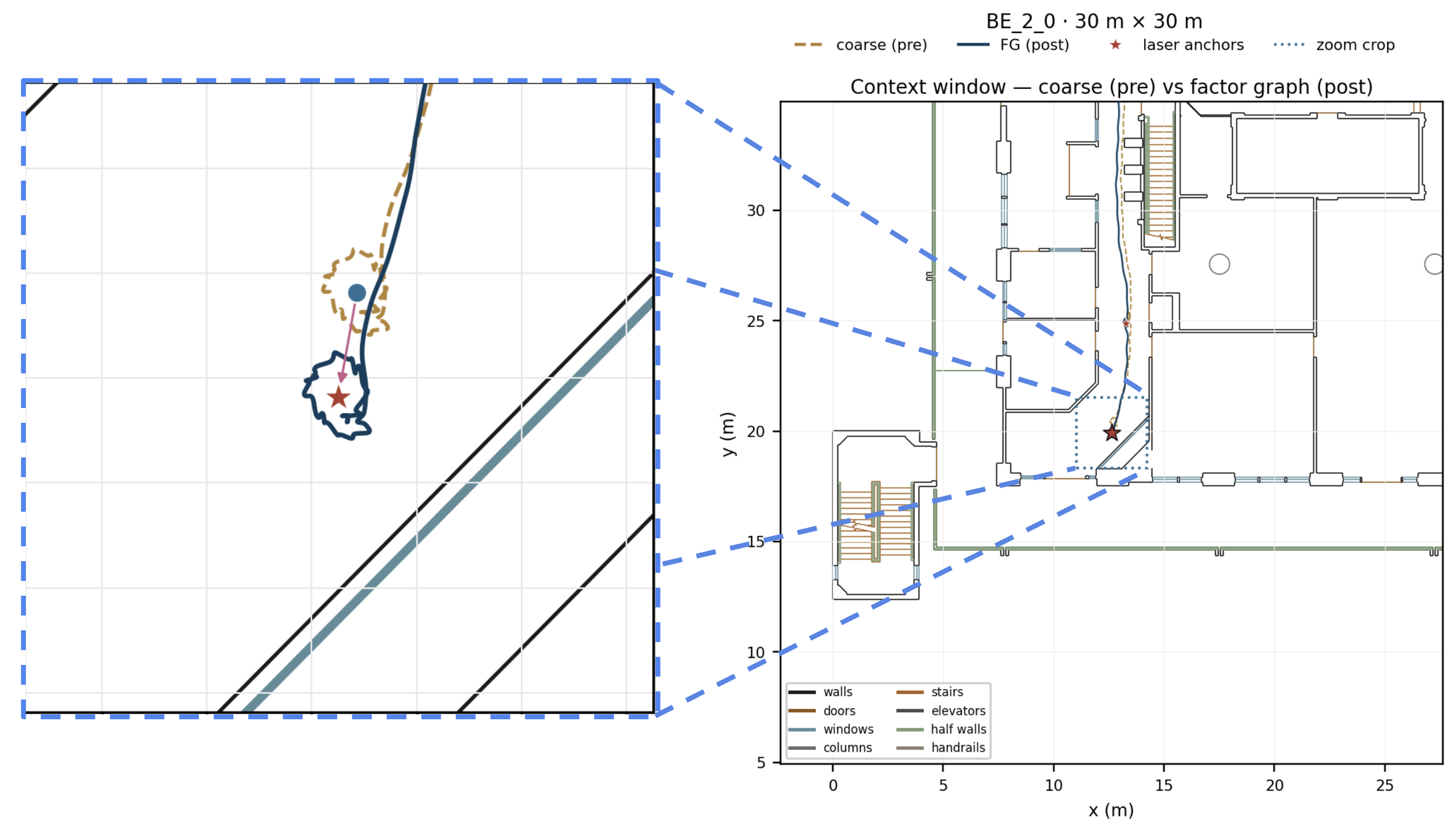}
\caption{Factor-graph alignment of an MPS trajectory into the floor plan frame using laser-anchored markers. Coarse trajectory (brown dashed line) is aligned by optimizing a factor graph, pulling each sweep center (blue dot) to agree with the corresponding anchors (red star), resulting in a trajectory that is aligned with the floor plan (black solid line).}
\label{fig:fg-alignment}
\end{figure}

MPS provides a closed-loop trajectory from its SLAM pipeline, which we subsample to the native RGB rate of $30$\,Hz.
Each panoramic sweep is captured on a floor marker; we treat these marker locations as anchors for placing the trajectory on the map.
We first apply a coarse manual alignment so the trajectory sits on the floor plan, then refine it with a factor graph that adds one virtual pose per sweep, initialized at the mean of that sweep's poses.
The graph has three kinds of factors: between consecutive poses along the walk; for frames inside a sweep, between each pose and the virtual sweep-center pose (preserving the observed spin rather than collapsing frames onto the center); and a position prior that pulls the sweep center toward the laser-measured anchor, leaving the heading of that virtual pose free.
The solver updates only planar position and heading; height, pitch, and roll remain those of MPS (Fig.~\ref{fig:fg-alignment}).
On a leave-every-other-anchor check (six recordings, 77 held-out panoramic sweeps), the factor graph recovers unused laser markers to $0.15$\,m mean / $0.09$\,m median.
Released ground truth uses every marker; this check is interpolation error at unused nails, not the residual of the shipped trajectory.

\section{Benchmark Framework}
\label{sec:benchmark}
We release the full $30$\,Hz recordings and floor-plan-aligned trajectories for research use.
The protocol below defines the splits, sampling rates, and tasks under which we report results, so methods can be compared under one standard.
\subsection{Problem Formulation}
\label{sec:problem}
Given a floor plan $\mathcal{F}$ and one or more visual observations $\mathcal{O}$, the goal is to estimate a planar camera pose $\mathbf{p}=(x,y,\theta)\in\mathrm{SE}(2)$ in the floor plan reference frame. We evaluate three complementary observation settings:
\begin{enumerate}
    \item \textbf{Direct single-frame localization:} one walking-frame RGB image, without temporal context.
    \item \textbf{Direct sweep localization:} multiple images acquired from approximately the same location during a near-stationary horizontal sweep.
    \item \textbf{Sequential localization:} a temporally ordered image stream with relative odometry. Relative motion is derived from the aligned trajectory as $\Delta\mathbf{T}_{t}=\mathbf{T}_{t-1}^{-1}\mathbf{T}_{t}$.
\end{enumerate}
These tasks isolate a single local view, increased angular coverage at a fixed location, and temporal accumulation along a trajectory. Prior work usually reports only the regime matching its design (snapshot, fixed-location scan as in PALMS/PALMS+~\cite{Cheng2024PALMSPA,Cheng2025PALMSMI}, or sequential filtering~\cite{Chen2024F3LocFA,West2025UnLocLD}). SlugTrails evaluates all three on the same buildings and ground truth because they reflect distinct practical ways to gather geometry under a limited FOV: one egocentric frame; a near-stationary horizontal sweep (``stop and look around''); and a walking stream that tests odometry-backed temporal fusion (Fig.~\ref{fig:thumbnail}).

\begin{table}[t]
\centering
\caption{Native sensing contracts.}
\label{tab:method-native}
\setlength{\tabcolsep}{3pt}
\footnotesize
\begin{tabular}{@{}lcccccc@{}}
\toprule
Config & HFOV & Rays & Step & Depth & Temporal & Train \\
\midrule
F$^{3}$Loc & $106^{\circ}$ & $11$ & $10^{\circ}$ & $10$\,m & filter & Gibson \\
UnLoc & $48.5^{\circ}$ & $10$ & $5^{\circ}$ & $\infty$ & filter & LaMAR \\
DisCo & $80^{\circ}$ & $9$ & $10^{\circ}$ & $20$\,m & filter & S3D \\
SemRayLoc & $80^{\circ}$ & $9$ & $10^{\circ}$ & $15$\,m & none & S3D \\
PP & --- & --- & --- & --- & CES/PF & --- \\
\bottomrule
\end{tabular}\\[2pt]
{\scriptsize Kept at evaluation; not a matched bakeoff. S3D\,=\,Structured3D. Temporal is the \emph{published} design---our sweep and sequential runs share one histogram filter. F$^{3}$Loc targets $106^{\circ}$ but Aria supplies $\approx 98^{\circ}$ (padded); PP consumes MoGe-2 depth rather than rays, it merges per-frame predictions for a single fire for sweep setting, and uses a particle filter for sequential merging.}
\end{table}

\paragraph{Train, validation, and test splits}
Because the benchmark includes fine-tuning, we split the six recordings into training, validation, and test sets.
Training and validation use four sessions (\texttt{BE\_0\_0}, \texttt{BE\_1\_0}, \texttt{E2\_2\_0}, \texttt{E2\_3\_0}).
From the time-ordered $30$\,Hz aligned trajectory of each session we take the middle $0.8$ as training and the first $0.1$ plus last $0.1$ as validation, so validation frames sit at the ends of the walk rather than being interleaved with the training samples.
We then keep every third frame ($10$\,Hz), giving $21012$ training frames and $5254$ validation frames.
Validation is used only for early stopping on ray loss; it is not a localization holdout.

Leave-out testing uses two recordings: in-distribution (ID) \texttt{BE\_2\_0}, a different floor of a building seen in training, and out-of-distribution (OOD) \texttt{PS\_2\_0}, a held-out building.
We evaluate all three tasks on these recordings.

\paragraph{Test sampling}
Single-frame and sequential samples come from walking intervals; sweeps come from annotated stationary rotations.
For single-frame localization we subsample walking video to $1$\,Hz, drop panoramic sweeps, discard frames whose absolute pitch exceeds $45^{\circ}$, and keep a pose only if it differs from previously kept samples by at least $1$\,m in position or $30^{\circ}$ in heading ($433$ frames).
For sweeps we sample the native $30$\,Hz stream along each panoramic spin and form multi-view sets of $3$, $4$, or $5$ images with union field of view $180^{\circ}$, $270^{\circ}$, or $360^{\circ}$, starting at $45^{\circ}$ intervals around the spin ($1008$ sweeps).
For sequential localization we subsample walking video to $2$\,Hz, drop panoramic sweeps, and draw fixed-length streams of $15$, $20$, $35$, $50$, or $100$ frames, $20$ sequences per length per recording ($200$ sequences).

\paragraph{Openings, glass, and half-walls}
Doors and windows are ambiguous for egocentric ray predictors: a door may be open or closed, and glass may be solid or see-through.
During SlugTrails fine-tuning, we do not supervise on door or window rays, so we do not risk giving the model a wrong or confusing depth, and we leave its prior training on openings intact.
At evaluation, each ray that meets a door or a window is compared to two ground-truth depths: the range to the opening surface if it is closed, and the range through the opening if it is open.
We take the ground truth closer to the predicted ray, so the method is not penalized for either valid hypothesis.

\subsection{Benchmarking Philosophy and Metrics}
\label{sec:protocol}
Our goal is to evaluate a dataset, not to propose a method.
We therefore run every system in its official released configuration (Table~\ref{tab:method-native}), as UnLoc~\cite{West2025UnLocLD} does for its comparisons, and we do not attempt a controlled architecture comparison: absolute cross-method numbers are reported for practitioners, but every claim we draw is \emph{within} a method, where architecture, preprocessing, and prior training data are held fixed and only SlugTrails varies.
That design is what makes the answers interpretable, and it lets us ask three questions:
(1)~how well do these models perform on SlugTrails off-the-shelf, and does SlugTrails fine-tuning help under that configuration, on single-frame, sweep, and sequential tasks;
(2)~does that gain generalize to another large indoor dataset, or are the models only learning about SlugTrails metadata (Aria appearance, floor plan style, and buildings);
(3)~if we match output geometry at inference without changing the network, how much of the difference between methods is FOV, depth range, and ray count.

We evaluate PALMS+ (PP)~\cite{Cheng2025PALMSMI}, F$^{3}$Loc~\cite{Chen2024F3LocFA}, UnLoc~\cite{West2025UnLocLD}, SemRayLoc~\cite{Grader2025SuperchargingFL}, and DisCo-FLoc~\cite{Zhong2026DisCoFLocSF} (written DisCo in tables).
PALMS+ is training-free (MoGe-2~\cite{Wang2025MoGe2AM} depth); DisCo-FLoc \textbf{FT} is the geometric RRP localizer.
We preprocess SlugTrails RGB to each method's native input: the ray methods (F$^{3}$Loc, UnLoc, DisCo-FLoc, SemRayLoc) share one gravity rectification followed by a FOV crop or pad onto that method's canvas and resolution, while PALMS+ keeps full Aria RGB-D at the measured headset FOV ($\sim$98$^\circ$).

\paragraph{Fine-tuning recipe}
Trainable methods start from stock checkpoints and keep their official training loss and depth-head range---masked L1 plus a shape term for F$^{3}$Loc and SemRayLoc, masked L1 for the DisCo-FLoc RRP decoder, and Laplace negative log-likelihood for UnLoc---with our only change being that door and window rays carry weight zero (Sec.~\ref{sec:problem}).
Each method also keeps its official frozen/trainable split: F$^{3}$Loc and SemRayLoc train the full encoder, DisCo-FLoc keeps DINO frozen, UnLoc keeps DA-V2 frozen.
We train with AdamW at $1{\times}10^{-4}$, cosine decay with $2$ warmup epochs, gradient clip $1.0$, and no AMP, on the four-session train carve, and early-stop on the validation carve ($100$ epochs, patience $15$; UnLoc $50$/$10$).
We evaluate the early-stopped best checkpoint, not the last epoch.
PALMS+ is training-free and is not fine-tuned.
Unless an ablation says otherwise, candidate poses use the circulation space mask.

\paragraph{Histogram filter for sweep and sequential}
Single-frame localization scores each frame independently.
Sweep and sequential localization fuse a sequence of observation likelihoods with a histogram filter: a motion update predicts the belief, which is then multiplied by the new observation.
All ray methods share one motion update---we used UnLoc's histogram filter setting from their LaMAR-HGE experiment: $\sigma_{x}=\sigma_{y}=\sigma_{\theta}=0.1$ with kernel windows $71$---driven by the full relative ego pose $(\Delta \mathrm{fwd}, \Delta \mathrm{lat}, \Delta\theta)$, so fusion is identical across methods while observation models stay native (Table~\ref{tab:method-native}).
SemRayLoc has no sequential filter in its published form; we graft the same one.
PALMS+ sweep localization is layout matching over the multi-view set rather than this filter, and PALMS+ sequential share the same setting.

We report R@1m30$^{\circ}$, R@1m, and R@5m.
Sweep uses last-frame ground truth; sequential uses the sequence-final pose.

\section{Experiments}
\label{sec:experiments}

\subsection{Experiment 1: Stock versus SlugTrails Fine-Tuning}
\label{sec:exp-adapt}
This experiment compares official released weights (\textbf{stock}) against initialization from those weights followed by SlugTrails fine-tuning (\textbf{FT}) on the leave-out split (ID=\texttt{BE\_2\_0}, OOD=\texttt{PS\_2\_0}).
PALMS+ has no neural weights, so its two settings are the official knobs (\textbf{default}) and CES knobs calibrated on the train carve (\textbf{calibrated}).
The same checkpoints are then scored under all three observation protocols: single-frame (Table~\ref{tab:main-adapt}), sweep (Table~\ref{tab:sweep}), and sequential (Table~\ref{tab:seq}).

\begin{table}[t]
\centering
\caption{Single-frame leave-out.}
\label{tab:main-adapt}
\setlength{\tabcolsep}{2.5pt}
\footnotesize
\begin{tabular}{@{}lccc@{\hskip 7pt}ccc@{}}
\toprule
& \multicolumn{3}{c}{ID} & \multicolumn{3}{c}{OOD} \\
\cmidrule(lr){2-4}\cmidrule(lr){5-7}
Recall@ & 1m30$^{\circ}$ & 1\,m & 5\,m & 1m30$^{\circ}$ & 1\,m & 5\,m \\
\midrule
UnLoc stock & 0.0$^\dagger$ & 0.0 & 0.034 & 0.0 & 0.01 & 0.111 \\
F$^{3}$Loc stock & 0.0 & 0.004 & 0.017 & 0.0 & 0.0 & 0.06 \\
SemRayLoc stock & 0.0 & 0.004 & 0.03 & 0.005 & 0.005 & 0.075 \\
DisCo stock & 0.004 & 0.004 & 0.034 & 0.01 & 0.01 & 0.116 \\
PP default & 0.004 & 0.073 & 0.12 & 0.0 & 0.025 & 0.095 \\
\midrule
SemRayLoc FT & 0.009 & 0.009 & 0.064 & 0.01 & 0.015 & 0.101 \\
PP calibrated & 0.009 & 0.043 & 0.077 & 0.0 & 0.05 & 0.131 \\
UnLoc FT & 0.026$^\dagger$ & 0.034 & 0.12 & 0.03 & 0.045 & \textbf{0.156} \\
DisCo FT & 0.06 & 0.064 & 0.107 & 0.04 & 0.05 & \textbf{0.156} \\
F$^{3}$Loc FT & \textbf{0.141} & \textbf{0.141} & \textbf{0.184} & \textbf{0.055} & \textbf{0.055} & 0.151 \\
\bottomrule
\end{tabular}\\[2pt]
{\scriptsize Columns: R@1m30$^{\circ}$ (primary), R@1m, R@5m; ID\,=\,\texttt{BE\_2\_0}, OOD\,=\,\texttt{PS\_2\_0}. Top block stock (PP: default), bottom block SlugTrails-FT (PP: calibrated); rows sorted by ID 1m30$^{\circ}$, best last. Tables~\ref{tab:sweep}--\ref{tab:ray-interp} follow this layout.}
\end{table}

\begin{table}[t]
\centering
\caption{Sweep leave-out (last-frame GT).}
\label{tab:sweep}
\setlength{\tabcolsep}{2.5pt}
\footnotesize
\begin{tabular}{@{}lccc@{\hskip 7pt}ccc@{}}
\toprule
& \multicolumn{3}{c}{ID} & \multicolumn{3}{c}{OOD} \\
\cmidrule(lr){2-4}\cmidrule(lr){5-7}
Recall@ & 1m30$^{\circ}$ & 1\,m & 5\,m & 1m30$^{\circ}$ & 1\,m & 5\,m \\
\midrule
F$^{3}$Loc stock & 0.0 & 0.0 & 0.019 & 0.0 & 0.003 & 0.025 \\
SemRayLoc stock & 0.007 & 0.012 & 0.027 & 0.013 & 0.017 & 0.094 \\
UnLoc stock & 0.024$^\dagger$ & 0.039 & 0.092 & 0.022 & 0.027 & 0.118 \\
DisCo stock & 0.036 & 0.065 & 0.147 & 0.066 & 0.091 & 0.17 \\
PP default$^\ast$ & 0.251 & 0.266 & 0.309 & 0.14 & 0.155 & 0.246 \\
\midrule
SemRayLoc FT & 0.019 & 0.022 & 0.068 & 0.037 & 0.051 & 0.17 \\
UnLoc FT & 0.034$^\dagger$ & 0.051 & 0.092 & 0.025 & 0.037 & 0.146 \\
DisCo FT & 0.159 & 0.174 & 0.246 & 0.131 & 0.138 & 0.219 \\
PP calibrated$^\ast$ & 0.193 & 0.21 & 0.256 & 0.163 & 0.168 & 0.251 \\
F$^{3}$Loc FT & \textbf{0.379} & \textbf{0.384} & \textbf{0.406} & \textbf{0.221} & \textbf{0.229} & \textbf{0.333} \\
\bottomrule
\end{tabular}\\[2pt]
{\scriptsize $^\ast$\,PP depth scaling: ground-overlap (default), none (calibrated). $^\dagger$\,See Sec.~\ref{sec:ablations}.}
\end{table}

\begin{table}[t]
\centering
\caption{Sequential leave-out (sequence-final pose).}
\label{tab:seq}
\setlength{\tabcolsep}{2.5pt}
\footnotesize
\begin{tabular}{@{}lccc@{\hskip 7pt}ccc@{}}
\toprule
& \multicolumn{3}{c}{ID} & \multicolumn{3}{c}{OOD} \\
\cmidrule(lr){2-4}\cmidrule(lr){5-7}
Recall@ & 1m30$^{\circ}$ & 1\,m & 5\,m & 1m30$^{\circ}$ & 1\,m & 5\,m \\
\midrule
UnLoc stock & 0.0$^\dagger$ & 0.01 & 0.06 & 0.09 & 0.1 & 0.24 \\
F$^{3}$Loc stock & 0.03 & 0.09 & 0.12 & 0.02 & 0.03 & 0.08 \\
DisCo stock & 0.05 & 0.07 & 0.1 & 0.18 & 0.28 & 0.41 \\
PP default & 0.11 & 0.12 & 0.25 & 0.05 & 0.05 & 0.16 \\
SemRayLoc stock & 0.18 & 0.18 & 0.19 & 0.12 & 0.13 & 0.22 \\
\midrule
PP calibrated & 0.11 & 0.12 & 0.23 & 0.07 & 0.07 & 0.16 \\
UnLoc FT & 0.22$^\dagger$ & 0.22 & 0.26 & 0.27 & 0.29 & 0.43 \\
SemRayLoc FT & 0.31 & 0.31 & 0.34 & 0.24 & 0.25 & 0.36 \\
DisCo FT & 0.33 & 0.49 & 0.52 & 0.29 & 0.4 & 0.48 \\
F$^{3}$Loc FT & \textbf{0.66} & \textbf{0.74} & \textbf{0.74} & \textbf{0.49} & \textbf{0.57} & \textbf{0.62} \\
\bottomrule
\end{tabular}\\[2pt]
{\scriptsize PP OOD: $28/100$ seq-finals collapse to the origin. $^\dagger$\,See Sec.~\ref{sec:ablations}.}
\end{table}

\paragraph{General observations}
Stock checkpoints are near zero on walking single frames: every trainable method is at or below $0.004$ ID R@1m30$^{\circ}$ (Table~\ref{tab:main-adapt}).
Stock models trained on Gibson, Structured3D, or LaMAR do not generalize to SlugTrails as-is.
Fine-tuning helps every trainable family on all three tasks, and \textbf{FT}$>$\textbf{stock} holds in all twelve method--task pairs of Tables~\ref{tab:main-adapt}--\ref{tab:seq}.
The largest single-frame lift is F$^{3}$Loc ($0.0\rightarrow 0.141$ ID R@1m30$^{\circ}$), followed by DisCo-FLoc ($0.004\rightarrow 0.06$) and UnLoc ($0.0\rightarrow 0.026$), while SemRayLoc stays weak ($0.0\rightarrow 0.009$).
Three patterns organize the rest.

\paragraph{Pattern 1---methods peak in the regime they were designed for}
A single walking frame is a narrow slice of a large floor, and each method's accumulator was written for a particular way of gathering that view.
PALMS+ stays near zero on singles ($0.004$ ID R@1m30$^{\circ}$) and jumps to $0.251$ on a stationary sweep---``stop and look around'' is the setting its layout matcher targets.
It does \emph{not} carry that improvement to sequential experiment ($0.11$, Table~\ref{tab:seq}).
The ray methods show the mirror image.
UnLoc barely moves on sweeps ($0.024\rightarrow 0.034$ stock$\rightarrow$FT, and $0.026\rightarrow 0.034$ single$\rightarrow$sweep for FT)---a spin at one marker does not repair a weak per-frame observation model---yet reaches $0.22$ once odometry and a filter are available, which is the regime its published protocol uses.
F$^{3}$Loc ships both multi-view and sequential inference, and it posts the highest absolute recall of any system we test in both regimes ($0.379$ sweep, $0.66$ sequential).

\paragraph{Pattern 2---gains compound across observation regimes}
The single-frame gain is not just added to by extra observations---it is multiplied by them.
F$^{3}$Loc FT goes $0.141\rightarrow 0.379\rightarrow 0.66$ across single, sweep, and sequential ($2.7\times$ then $4.7\times$ its single-frame recall), and DisCo-FLoc FT goes $0.06\rightarrow 0.159\rightarrow 0.33$ ($2.7\times$, $5.5\times$).
Their stock counterparts do not compound: F$^{3}$Loc stock is $0.0\rightarrow 0.0\rightarrow 0.03$ and DisCo-FLoc stock $0.004\rightarrow 0.036\rightarrow 0.05$.
The filter is the same in both cases, so the extra observations are not creating information---they are propagating probability mass that a well-adapted observation model already places near the true pose.
Fine-tuning raises per-frame posterior quality, and multi-view coverage and temporal fusion then keep amplifying it; a stock model gives the filter little correct mass to carry forward.

\paragraph{Pattern 3---building difference shifts the level, not the trend}
ID is a held-out floor of a building seen in training (\texttt{BE\_2\_0}); OOD is an unseen building (\texttt{PS\_2\_0}), and it is also harder to localize within due to its large open spaces.
Recall is accordingly lower on OOD---F$^{3}$Loc FT $0.141\rightarrow 0.055$ and DisCo-FLoc FT $0.06\rightarrow 0.04$ on single frames---but the growth pattern is identical on both: fine-tuning beats stock, and more frames beat fewer (F$^{3}$Loc FT OOD $0.055\rightarrow 0.221\rightarrow 0.49$ across single, sweep, and sequential).
The unseen building lowers the absolute level without changing what helps.

\paragraph{Exceptions}
SemRayLoc does not fit the compounding story: it is the weakest fine-tuned system on singles ($0.009$) yet reaches $0.31$ sequential, above UnLoc's $0.22$ in that regime despite a much lower single-frame score, suggesting its semantic observations help temporal fusion more than single-frame ranking.

\subsection{Experiment 2: Cross-dataset Generalization}
\label{sec:exp-lamar}
A possible objection is that fine-tuning only memorizes SlugTrails-specific metadata (Aria appearance, our floor plan layers, our buildings).
We therefore freeze the same stock and SlugTrails-FT weights and evaluate on LaMAR~\cite{Sarlin2022LaMARBL} CAB iOS queries, with \emph{no} LaMAR fine-tuning.
LaMAR does not release architectural floor plans; we first manually define per-floor boundaries to crop the video trajectories. Then, we hand-trace walls and circulation space masks from the provided point clouds and use those masks as pose support, matching the SlugTrails circulation protocol.
The CAB iOS query set spans $5$ floors, $34$ trajectories, and $\sim$3.0\,km of path length after pitch and outer-mask filters ($n{=}4031$ frames; $63$ floor segments).

This is a useful split of prior data: F$^{3}$Loc never trained on LaMAR, while UnLoc did.
If SlugTrails FT were only Aria and floor plan overfitting, we would not expect a gain on LaMAR, and especially not for UnLoc, which already saw this domain.
\begin{table}[t]
\centering
\caption{Cross-dataset generalization on LaMAR CAB.}
\label{tab:lamar-cab}
\setlength{\tabcolsep}{2.5pt}
\footnotesize
\begin{tabular}{@{}lccc@{\hskip 6pt}c@{}}
\toprule
& \multicolumn{3}{c}{Single-frame} & Sequential \\
\cmidrule(lr){2-4}\cmidrule(lr){5-5}
Recall@ & 1m30$^{\circ}$ & 1\,m & 5\,m & 1\,m \\
\midrule
F$^{3}$Loc stock & 0.0005 & 0.0025 & 0.039 & 0.048 \\
UnLoc stock & 0.0045$^\dagger$ & 0.0102 & \textbf{0.102} & 0.063 \\
\midrule
F$^{3}$Loc FT & 0.0057 & 0.0084 & 0.062 & \textbf{0.143} \\
UnLoc FT & \textbf{0.0089}$^\dagger$ & \textbf{0.0181} & 0.099 & 0.127 \\
\bottomrule
\end{tabular}\\[2pt]
{\scriptsize Same checkpoints as Table~\ref{tab:main-adapt}; no LaMAR fine-tuning. Single-frame: $n{=}4031$ iOS queries. Sequential: R@1m at the last pose after offline fusion over each of 63 iOS segments (sequence-final; not the same queries as the single-frame columns). $^\dagger$\,See Sec.~\ref{sec:ablations}.}
\end{table}

\paragraph{Observations}
SlugTrails FT improves both methods on CAB single frames (Table~\ref{tab:lamar-cab}): F$^{3}$Loc $0.0005\rightarrow 0.0057$ R@1m30$^{\circ}$ ($0.0025\rightarrow 0.0084$ R@1m), UnLoc $0.0045\rightarrow 0.0089$ ($0.0102\rightarrow 0.0181$ R@1m).
The sequential probe in the same table is larger: sequential R@1m UnLoc $0.063\rightarrow 0.127$, F$^{3}$Loc $0.048\rightarrow 0.143$.
Gains on a dataset F$^{3}$Loc never saw, and on a dataset UnLoc already trained on, support treating SlugTrails as useful large-building adaptation data.
One caveat is that absolute CAB recall stays well below UnLoc's published numbers, likely due to the lack of GT floor plan; we do not claim a LaMAR state-of-the-art reproduction (Sec.~\ref{sec:ablations}).

\subsection{Experiment 3: Matched Output Geometry}
\label{sec:exp-matched}
The methods differ in many ways at once---FOV, ray count, angular step, maximum depth, backbone, and training data (Table~\ref{tab:method-native})---so Experiment~1 cannot tell us whether the output-space parameters or the methods themselves drive the differences in performance.
We therefore run a simple probe: without tampering with any method's internals, we restrict its output space at matching time and observe how each parameter changes recall.
No network is retrained and no weights change; we only select, subsample, or clip the rays a method already predicts.
Each row of Table~\ref{tab:ray-interp} changes essentially one parameter: F$^{3}$Loc keeps the central $9$ of its $11$ rays (narrowing FOV span from $\pm 50^{\circ}$ to $\pm 40^{\circ}$), DisCo-FLoc and SemRayLoc clip only their depth consume range to $10$\,m, and UnLoc subsamples its angular step from $5^{\circ}$ to $10^{\circ}$ and clips range.
The probe can only remove observation content, never add it, so no method's FOV can be widened.

\paragraph{Field of view and ray span}
Narrowing FOV costs recall, and the cost is not small.
F$^{3}$Loc is the only method wide enough to narrow toward $80^{\circ}$, and dropping its outer ray pair takes it from $0.141$ to $0.115$ ID and $0.055$ to $0.04$ OOD R@1m30$^{\circ}$---roughly a fifth to a quarter of its recall for $20^{\circ}$ of coverage.
Those outer rays see down the corridor and across intersections, exactly where a floor plan is locally ambiguous, so F$^{3}$Loc's high single-frame recall in Experiment~1 rests on usable geometry rather than on a more permissive configuration.

\paragraph{Maximum depth}
Depth range is the one parameter that matters more for large floors, and clipping it surprisingly helps.
DisCo-FLoc natively consumes rays out to $20$\,m; clipping to $10$\,m raises ID from $0.06$ to $0.077$ and more than doubles OOD, $0.04$ to $0.09$---the highest OOD recall of any configuration in the table.
Long predicted rays cross open space and windows where monocular depth is least reliable, so trusting them injects wrong evidence into the likelihood; part of DisCo-FLoc's Experiment~1 deficit is this parameter, not its architecture.
The effect is not universal---SemRayLoc clipped from $15$\,m to $10$\,m is flat on ID ($0.013$) and worse on OOD ($0.01\rightarrow 0.005$)---which suggests clipping helps only when the underlying long-range predictions were being trusted and were wrong.

\paragraph{Angular resolution}
At fixed FOV, ray density matters, and it matters most for coarse localization.
Halving UnLoc's ray count by moving from a $5^{\circ}$ to a $10^{\circ}$ step costs little at the strict threshold ($0.026\rightarrow 0.021$ ID R@1m30$^{\circ}$) but nearly halves R@5m ($0.12\rightarrow 0.064$).
Dense rays over a narrow FOV appear to be what keeps UnLoc's posterior in the right region of the floor at all, and its fine angular step is partly compensating for the coverage it does not have.

\paragraph{What the probe does not settle}
This is the closest approximation to a matched comparison that stock weights allow, but it is not a controlled one: weights, preprocessing, and training datasets still differ, and each method was trained to consume its native geometry.

\begin{table}[t]
\centering
\caption{Post-hoc ray interpretation: config edits and recall change.}
\label{tab:ray-interp}
\setlength{\tabcolsep}{2.5pt}
\footnotesize
\begin{tabular}{@{}lcccc@{}}
\toprule
Method & HFOV & Rays & Step & Depth \\
\midrule
SemRayLoc$^\ddagger$ & $80^{\circ}$ & $9$ & $10^{\circ}$ & \textbf{$15$\,m$\rightarrow$$10$\,m} \\
UnLoc$^{\S}$ & $48.5^{\circ}$ & $10$ & \textbf{$5^{\circ}$$\rightarrow$$10^{\circ}$} & \textbf{$\infty$$\rightarrow$$10$\,m} \\
DisCo & $80^{\circ}$ & $9$ & $10^{\circ}$ & \textbf{$20$\,m$\rightarrow$$10$\,m} \\
F$^{3}$Loc & \textbf{$106^{\circ}$$\rightarrow$$\sim$80$^{\circ}$} & \textbf{$11$$\rightarrow$$9$} & $10^{\circ}$ & $10$\,m \\
\bottomrule
\end{tabular}\\[4pt]
\begin{tabular}{@{}lccc@{\hskip 7pt}ccc@{}}
\toprule
& \multicolumn{3}{c}{ID $\Delta$} & \multicolumn{3}{c}{OOD $\Delta$} \\
\cmidrule(lr){2-4}\cmidrule(lr){5-7}
Recall@ & 1m30$^{\circ}$ & 1\,m & 5\,m & 1m30$^{\circ}$ & 1\,m & 5\,m \\
\midrule
SemRayLoc$^\ddagger$ & --- & --- & \textcolor{red}{$-0.021$} & \textcolor{red}{$-0.005$} & \textcolor{red}{$-0.015$} & \textcolor{red}{$-0.041$} \\
UnLoc$^{\S\dagger}$ & \textcolor{red}{$-0.005$} & \textcolor{red}{$-0.004$} & \textcolor{red}{$-0.056$} & \textcolor{red}{$-0.005$} & \textcolor{red}{$-0.015$} & \textcolor{red}{$-0.030$} \\
DisCo & \textcolor{green}{$+0.017$} & \textcolor{green}{$+0.017$} & \textcolor{red}{$-0.004$} & \textcolor{green}{$+0.050$} & \textcolor{green}{$+0.051$} & \textcolor{green}{$+0.020$} \\
F$^{3}$Loc & \textcolor{red}{$-0.026$} & \textcolor{red}{$-0.026$} & \textcolor{red}{$-0.026$} & \textcolor{red}{$-0.015$} & \textcolor{red}{$-0.010$} & \textcolor{red}{$-0.035$} \\
\bottomrule
\end{tabular}\\[2pt]
{\scriptsize All FT checkpoints (Table~\ref{tab:main-adapt}); no retraining. Top: native value, or \textbf{native$\rightarrow$matched} where edited. Bottom: matched minus native (--- = no change). $^{\S}$\,UnLoc cannot be widened---step subsample + range clip only. $^\ddagger$\,Earlier checkpoint than Table~\ref{tab:main-adapt}. $^\dagger$\,See Sec.~\ref{sec:ablations}.}
\end{table}

\subsection{Ablation Studies and Open Anomalies}
\label{sec:ablations}

\paragraph{Train-from-scratch versus fine-tuning}
To separate ``SlugTrails data helps'' from ``SlugTrails data is sufficient,'' we also train-from-scratch on SlugTrails---same data, same schedule, same losses as the FT settings, with only the initialization removed (ImageNet backbone for F$^{3}$Loc, foundation backbone only for DisCo-FLoc).
Train-from-scratch stays far below FT on the same leave-out: F$^{3}$Loc $0.009$ vs.\ $0.141$ ID R@1m30$^{\circ}$ ($0.015$ vs.\ $0.055$ OOD), and DisCo-FLoc $0.0$ vs.\ $0.06$ ID ($0.0$ vs.\ $0.04$ OOD).
Since SlugTrails is held fixed across the two settings, the gap is attributable to the prior training datasets---Gibson and Structured3D---being present or absent.
Those datasets are household-scale and synthetic, and checkpoints trained on them alone are near zero on SlugTrails (Table~\ref{tab:main-adapt}), yet removing them costs significant recall.

Read across all three regimes, recall increases monotonically with the amount of indoor data a model has seen: household-scale pretraining alone is near zero, SlugTrails alone is near zero, and the two together are the best result at every observation setting.
We see no sign of saturation at either end of that progression.
The practical implication is that the largest bottleneck for floor-plan localization is data rather than architecture: methods improve most when more indoor data---especially large buildings with heavy ambiguity, which remain scarce---is added to what they already have.

\paragraph{Outline versus circulation pose support}
On \emph{single-frame} leave-out only, we compare circulation pose support (main tables) to searching the full floor-plan outline.
Expanding to the outline barely changes ID but hurts OOD: F$^{3}$Loc FT $0.141\rightarrow 0.132$ ID R@1m30$^{\circ}$ vs.\ $0.055\rightarrow 0.01$ OOD; DisCo-FLoc FT $0.06\rightarrow 0.051$ ID vs.\ $0.04\rightarrow 0.005$ OOD.

\section{Conclusion}
\label{sec:conclusion}
\label{sec:discussion}
SlugTrails studies floor plan localization under realistic large-building egocentric sensing, with a single protocol spanning single-frame, multi-view sweep, and sequential observations.
Our first experiment shows that stock state-of-the-art methods generalize poorly to large academic floors---every trainable family is at or below $0.004$ ID R@1m30$^{\circ}$ on walking single frames---and that fine-tuning on SlugTrails helps all of them, on all three tasks (F$^{3}$Loc $0.0\rightarrow 0.141$ single, $0.0\rightarrow 0.379$ sweep, $0.03\rightarrow 0.66$ sequential).
Three patterns recur: methods peak in the observation regime they were designed for; adaptation gains compound with additional observations, since temporal fusion can only propagate probability mass that a well-adapted observation model already places near the true pose; and an unseen building lowers absolute recall without changing which interventions help.
Our second experiment tests cross-dataset generalization: freezing the same checkpoints and evaluating on LaMAR with hand-traced floor plans, SlugTrails fine-tuning improves both a method that never trained on LaMAR (F$^{3}$Loc, sequential R@1m $0.048\rightarrow 0.143$) and one that did (UnLoc, $0.063\rightarrow 0.127$), so the adaptation generalizes to another large indoor dataset rather than only fitting our buildings.

Smaller experiments sharpen two points.
First, train-from-scratch on SlugTrails alone stays far below fine-tuning from stock weights ($0.009$ vs.\ $0.141$ for F$^{3}$Loc), even though those stock weights come from household-scale and synthetic interiors.
Recall thus appears to increase monotonically with the total amount of indoor data a model has seen, with no sign of saturation at either end---which we read as the central practical finding: the bottleneck for floor-plan localization is data, not architecture, and the most direct path forward is collecting more large-building indoor data.
Second, restricting each method's output space post-hoc shows that some published sensing choices are simply mismatched to large floors: clipping DisCo-FLoc's depth consume range from $20$\,m to $10$\,m more than doubles its OOD recall ($0.04\rightarrow 0.09$), whereas narrowing F$^{3}$Loc's field of view costs recall.
Together these results indicate that SlugTrails is useful on both sides of the benchmark: as evaluation data that separates methods and exposes regime-specific and out-of-distribution failures, and as training data that measurably improves stock systems, including beyond its own domain.

Several limitations bound these claims.
SlugTrails covers a limited number of academic buildings compared with generic vision datasets; its value lies in combining verified floor plans---which are hard to obtain at all, given privacy and building-security constraints---with realistic egocentric trajectories and controlled evaluation in large spaces with heavy circulation.
Because we evaluate stock (official released) configurations rather than a unified retraining protocol (Sec.~\ref{sec:protocol}), our absolute cross-method numbers are observations rather than a controlled comparison.
Finally, our ground truth comes from laser-surveyed anchors fused with a factor graph rather than a dedicated survey-grade SLAM or laser-scanning pipeline: leaving out every other anchor recovers unused markers to $0.15$\,m mean and $0.09$\,m median error, so SlugTrails poses are consistent with the floor plan but not centimeter-accurate in the sense of SLAM benchmarks, and sub-meter comparisons should be interpreted with that in mind.
We release the dataset, protocols, and tools to help evaluate and improve floor plan localization for realistic indoor deployment.

\section*{Acknowledgment}
We thank Kenny Murray, Parker Welch, and Max Tcheng of UCSC PPDO, as well as others in PPDO, for sharing the floor plans and supporting this research.
We thank Brice Chin and Matthew Huang for their contributions in constructing this dataset.
The authors used Cursor to assist with: (1)~polishing and refining author-drafted text across all sections; (2)~generating and debugging experiment and evaluation code; and (3)~creating illustrative figures and diagrams. In all cases, the authors provided the draft text, structural outlines, and experiment results; the AI tool was used for refinement and implementation. All AI-generated content underwent multiple rounds of human review and editing. The authors take full responsibility for the content of this publication.

\bibliographystyle{IEEEtran}
\bibliography{references}

\end{document}